\documentclass[11pt,a4paper]{article}
\usepackage[hyperref]{acl2021}
\usepackage{times}
\usepackage{latexsym}

\usepackage{microtype}

\usepackage{graphicx}
\usepackage{todonotes}

\usepackage{etoolbox,siunitx}
\usepackage{multirow}
\usepackage{amsmath}
\usepackage{booktabs}

\usepackage{etoolbox}

\aclfinalcopy %
\def\aclpaperid{651} %

\newcommand{\squishlist}{
 \begin{list}{$\bullet$}
  { \setlength{\itemsep}{0pt}
     \setlength{\parsep}{3pt}
     \setlength{\topsep}{3pt}
     \setlength{\partopsep}{0pt}
     \setlength{\leftmargin}{1.5em}
     \setlength{\labelwidth}{1em}
     \setlength{\labelsep}{0.5em} } }

\newcommand{\squishend}{
  \end{list}  }

\newcommand\ignore[1]{}
  
\title{Coreference Reasoning in Machine Reading Comprehension}

\author{Mingzhu Wu$^1$, Nafise Sadat Moosavi$^1$, Dan Roth$^2$, Iryna Gurevych$^1$\\ \\ $^1$UKP Lab, Technische Universit¨at Darmstadt\\ $^2$Department of Computer and Information Science, UPenn\\  $^1$\url{https://www.ukp.tu-darmstadt.de} \\ $^2$  \url{https://www.seas.upenn.edu/\~danroth/}}

\date{}

\begin{document}
\maketitle
\begin{abstract}
Coreference resolution is essential for natural language understanding and has been long studied in NLP. In recent years, as the format of Question Answering (QA) became a standard for machine reading comprehension (MRC), there have been data collection efforts, e.g., \citet{dasigi2019quoref}, that attempt to evaluate the ability of MRC models to reason about coreference. However, as we show, coreference reasoning in MRC is a greater challenge than earlier thought; MRC datasets do not reflect the natural distribution and, consequently, the challenges of coreference reasoning. Specifically, success on these datasets does not reflect a model's proficiency in  coreference reasoning. 
We propose a methodology for  creating MRC datasets that better reflect the challenges of coreference reasoning and use it to create a sample evaluation set. The results on our dataset show that state-of-the-art models still struggle with these phenomena. Furthermore, we develop an effective way to use naturally occurring coreference phenomena from existing coreference resolution datasets when training MRC models. This allows us to show an improvement in the coreference reasoning abilities of state-of-the-art models.\footnote{The code and the resulting dataset are available at \url{https://github.com/UKPLab/coref-reasoning-in-qa}.}  
\end{abstract}

\section{Introduction}
Machine reading comprehension is the ability to read and understand the given passages and answer questions about them. 
Coreference resolution is the task of finding different 
expressions that refer to the same real-world entity.
The tasks of coreference resolution and machine reading comprehension have moved closer to each other.
Converting coreference-related datasets into an MRC format improves the performance on some coreference-related datasets \citep{wu-etal-2020-corefqa,aralikatte2020simple}.
There are also various datasets for the task of reading comprehension on which the model requires to perform coreference reasoning to answer some of the questions, e.g., DROP \cite{dua-etal-2019-drop}, DuoRC \cite{saha2018duorc}, MultiRC \cite{MultiRC2018}, etc.%

\begin{figure}[!htb]
\centering
\includegraphics[width=0.37\paperwidth]{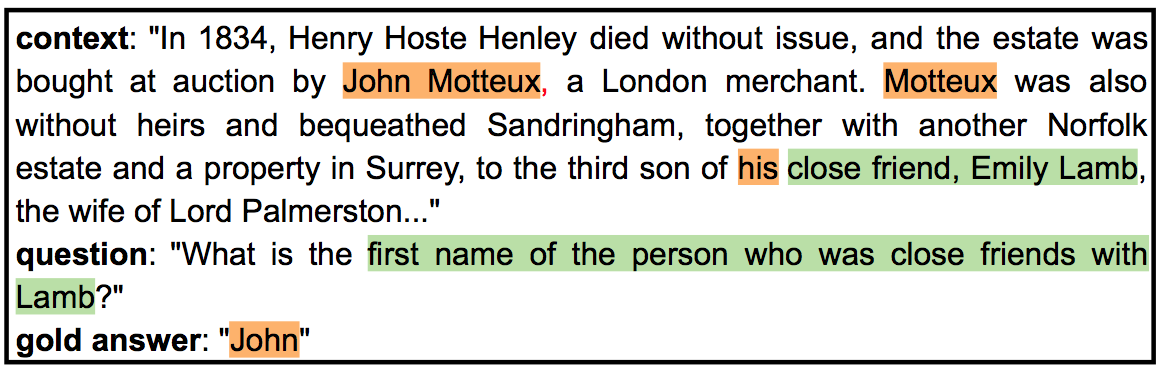}
\caption{A sample from the Quoref dataset. 
}
\label{coref_example}
\end{figure}

Quoref \citep{dasigi2019quoref} is a dataset that is particularly designed for evaluating coreference understanding of MRC models. 
Figure~\ref{coref_example} shows a QA sample from Quoref in which the model needs to resolve the coreference relation between ``his'' and ``John Motteux'' to answer the question.

Recent large pre-trained language models reached high performance on Quoref.
However, our results and analyses suggest that this dataset contains artifacts and does not reflect the natural distribution and, therefore, the challenges of coreference reasoning.
As a result, high performances on Quoref do not necessarily reflect the coreference reasoning capabilities of the examined models and answering questions that require coreference reasoning might be a greater challenge than current scores suggest.

In this paper, we propose two solutions to address this issue.
First, we propose a methodology for creating MRC datasets that better reflect the coreference reasoning challenge.
We release a sample challenging evaluation set containing 200 examples by asking an annotator to create new question-answer pairs using our methodology and based on existing passages in Quoref. 
We show that this dataset contains fewer annotation artifacts, and its distribution of biases is closer to a coreference resolution dataset.  
The performance of state-of-the-art models on Quoref considerably drops on our evaluation set suggesting that (1) coreference reasoning is still an open problem for MRC models, and (2) our methodology opens a promising direction to create future challenging MRC datasets.

Second, we propose to directly use coreference resolution datasets for training MRC models to improve their coreference reasoning.
We automatically create a question whose answer is a coreferring expression $m_1$
using the BART model \citep{lewis-etal-2020-bart}. We then consider this question, $m_1$'s antecedent, and the corresponding document as a new (question, answer, context) tuple.
This data helps the model learning to resolve the coreference relation between $m_1$ and its antecedent to answer the question. 
We show that incorporating these additional data improves the performance of the state-of-the-art models on our new evaluation set. %

Our main contributions are as follows:
\squishlist
\item We show that Quoref does not reflect the natural challenges of coreference reasoning and propose a methodology for creating MRC datasets that better reflect this challenge. 
\item We release a sample challenging dataset that is manually created by an annotator using our methodology. The results of state-of-the-art MRC models on our evaluation set show that, despite the high performance of MRC models on Quoref, answering questions based on coreference reasoning is still an open challenge. 
\item We propose an approach to use existing coreference resolution datasets for training MRC models. We show that, while coreference resolution and MRC datasets are independent and belong to different domains, our approach improves the coreference reasoning of state-of-the-art MRC models. 
\squishend

\section{Related Work}

\subsection{Artifacts in NLP datasets}
One of the known drawbacks of many NLP datasets is that they contain artifacts.\footnote{I.e., the conditional distribution of the target label based on specific attributes of the training domain diverges while testing on other domains.}
Models tend to exploit these easy-to-learn patterns in the early stages of training \citep{arpit2017closer,liu2020early,utama-etal-2020-towards}, and therefore, they may not focus on learning harder patterns of the data that are useful for solving the underlying task.
As a result, overfitting to dataset-specific artifacts limits the robustness and generalization of NLP models.

There are two general approaches to tackle such artifacts: (1) adversarial filtering of biased examples, i.e., examples that contain artifacts, and (2) debiasing methods.
In the first approach, potentially biased examples are discarded from the dataset, either after the dataset creation \citep{zellers-etal-2018-swag, yang-etal-2018-hotpotqa,le2020adversarial,Bartolo_2020}, or while creating the dataset \citep{dua-etal-2019-drop, chen2019codah,nie-etal-2020-adversarial}. 

In the second approach, they first recognize examples that contain artifacts, and use this knowledge in the training objective to either skip or down-weight biased examples \citep{he-etal-2019-unlearn,clark-etal-2019-dont}, or to regularize the confidence of the model on those examples \citep{utama-etal-2020-mind}.
The use of this information in the training objective improves the robustness of the model on adversarial datasets \citep{he-etal-2019-unlearn,clark-etal-2019-dont,utama-etal-2020-mind}, i.e., datasets that contain counterexamples in which relying on the bias results in an incorrect prediction.
In addition, it can also improve in-domain performances as well as generalization across various datasets that represent the same task \citep{wu2020improving,utama-etal-2020-towards}.  

While there is an emerging trend of including adversarial models in data collection, their effectiveness is not yet compared with using debiasing methods, e.g., whether they are still beneficial when we use debiasing methods or vice versa.

\subsection{Joint QA and Coreference Reasoning}
There are a few studies on the joint understanding of coreference relations and reading comprehension.
\citet{wu-etal-2020-corefqa} propose to formulate coreference resolution as a span-prediction task by generating a query for each mention using the surrounding context, thus converting coreference resolution to a reading comprehension problem. They leverage the plethora of existing MRC datasets for data augmentation and improve the generalization of the coreference model. 
In parallel to \citet{wu-etal-2020-corefqa}, \citet{aralikatte2020simple} also cast ellipsis and coreference resolution as reading comprehension tasks. They leverage the existing neural architectures designed for MRC for ellipsis resolution and outperform the previous best results.
In a similar direction, \citet{hou-2020-bridging} propose to cast bridging anaphora resolution as question answering and present a question answering framework for this task. 
However, none of the above works
investigate the impact of using coreference data on QA.

\citet{dua-etal-2020-benefits} use Amazon Mechanical Turkers to annotate the corresponding coreference chains of the answers in the passages of Quoref for 2,000 QA pairs. They then use this additional coreference annotation for training a model on Quoref. They show that including these additional coreference annotations improves the overall performance on Quoref.
The proposed method by \citet{dua-etal-2020-benefits} requires annotating additional coreference relations on every new coreference-aware QA dataset. Contrary to this, our approach uses existing coreference resolution datasets, and therefore, applies to any new QA dataset without introducing any additional cost.    
\section{How Well Quoref Presents Coreference Reasoning?}
\label{sect:quoref_bias}
For creating the Quoref dataset, annotators first identify coreferring expressions and then ask questions that connect the two coreferring expressions.
\citet{dasigi2019quoref} use a BERT-base model \citep{devlin2019bert} that is fine-tuned on the SQuAD dataset \cite{rajpurkar-etal-2016-squad} as an adversarial model to exclude QA samples that the adversarial model can already answer.
The goal of using this adversarial model is to avoid including question-answer pairs that can be solved using surface cues.
They claim that most examples in Quoref cannot be answered without coreference reasoning.

If we fine-tune a RoBERTa-large model on Quoref, it achieves 78 F1 score while the estimated human performance is around 93 F1 score \citep{dasigi2019quoref}.
This high performance, given that RoBERTa can only predict continuous span answers while Quoref also contains discontinuous answers, indicates that either (1) Quoref presents coreference-aware QA very well so that the model can properly learn coreference reasoning from the training data, (2) pretrained transformer-based models have already learned coreference reasoning during their pre-training, e.g., as suggested by \citet{tenney-etal-2019-bert} and \citet{clark-etal-2019-bert}, or (3) coreference reasoning is not necessarily required for solving most examples. 

\begin{figure}[!htb]
\centering
\includegraphics[width=0.37\paperwidth]{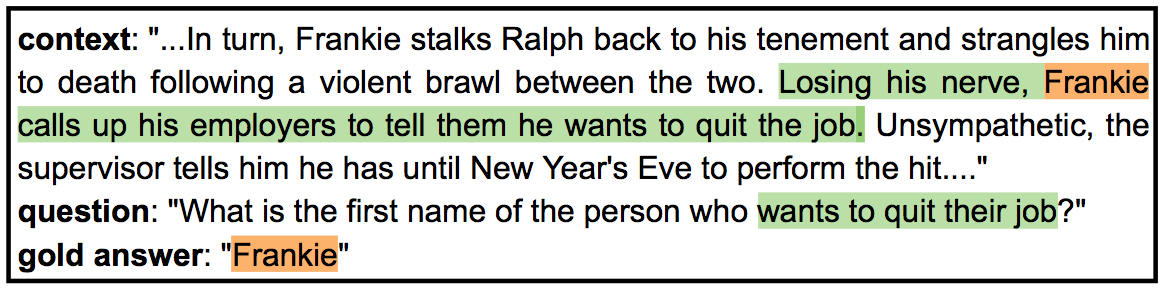}
\caption{A QA example that relies on simple lexical overlap without requiring coreference reasoning.}
\label{lexical_example}
\end{figure}

In this section, we investigate whether Quoref contains the known artifacts of QA datasets, and therefore, models can solve some of the QA pairs without performing coreference reasoning.
Figure \ref{lexical_example} shows such an example where simple lexical cues are enough to answer the question despite the fact that coreference expressions ``Frankie'' and ``his'' were included in the corresponding context.

We investigate five artifacts (biases) as follows: 

\squishlist
\item Random named entity: the majority of answers in Quoref are person names. To evaluate this artifact, we randomly select a PERSON named entity from the context as the answer.\footnote{We use spaCy \citep{honnibal-johnson:2015:EMNLP} for NER.}
\item Wh-word \citep{weissenborn-etal-2017-making}: to recognize the QA pairs that can be answered by only using the interrogative adverbs from the question, we train a model on a variation of the training dataset in which questions only contain interrogative adverbs.
\item Empty question \citep{sugawara2019assessing}: to recognize QA pairs that are answerable without considering the question,\footnote{E.g., this can indicate the bias of the model to select the most frequent named entity in the context as the answer.} we train a QA model only on the contexts and without questions.

\item Semantic overlap \citep{Jia_2017}: for this artifact, we report the ratio of the QA pairs whose answers lie in the sentence of the context that has the highest 
semantic
similarity to the question. We use sentence-BERT \citep{reimers-gurevych-2019-sentence} to find the most similar sentence.
\item Short distance reasoning: for this bias, we train a model only using the sentence of the context that is the most similar to the question, instead of the whole context. We exclude the question-answer pairs in which the most similar sentence does not contain the answer. This model will not learn to perform coreference reasoning when the related coreferring pairs are not in the same sentence.
\squishend

For wh-word, empty question, and short distance reasoning, we use the TASE model \cite{segal2020simple} to learn the bias.
Biased examples are then those that can be correctly solved by these models.
We only change the training data for biased example detection, if necessary, and the development set is unchanged.
The \emph{Quoref} column in Table~\ref{table:bias_results} reports the %
proportion of biased examples in the Quoref development set.

\begin{table}[!htb]
\footnotesize
\centering
\begin{tabular}{lrr}
\toprule
 \bfseries Bias  & \bfseries Quoref &  \bfseries CoNLL$_{bart}$  \\ 
\midrule
random named entity  & 9.39   & 1.52   \\
wh-word  &  22.99 & 13.12 \\ 
empty question  & 21.51 & 11.60    \\
semantic overlap & 28.66 & 21.38  \\
short-distance reasoning    & \bfseries 50.70 & 9.86 \\
\bottomrule
\end{tabular}
\caption{The proportion of examples in the Quoref development set and CoNLL-2012 coreference resolution dataset that contain each of the examined biases.}
\label{table:bias_results}
\end{table}

We also investigate whether these biases have similar ratios in a coreference resolution dataset.
We use the CoNLL-2012 coreference resolution dataset \citep{2012-ontonotes} and convert it to a reading comprehension format, i.e., CoNLL$_{bart}$ in Section~\ref{sect:method}.\footnote{We report the bias ratios of CoNLL$_{dec}$ in Section~\ref{sect:method} in the appendix.}
This data contains question-answer pairs in which the question is created based on a coreferring expression in CoNLL-2012, and the answer is its closest antecedent.
We split this data into training and test sets and train bias models on the training split.
The \emph{CoNLL$_{bart}$} column in Table~\ref{table:bias_results} shows the bias proportions on this data.

As we see, the short distance reasoning is the most prominent bias in the Quoref dataset. However, the ratio of such biased examples is only around 10\% in CoNLL-2012.
Therefore, apart from the examples that can be solved without coreference reasoning,\footnote{E.g., about 20\% of examples can be answered without considering the question.} the difficulty of the required coreference reasoning in the remaining examples is also not comparable with naturally occurring coreference relations in a coreference resolution dataset. 

As a result, high performance on Quoref does not necessarily indicate that the model is adept at performing coreference reasoning.
\begin{table*}[!htb]
\footnotesize
\centering
\begin{tabular}{l|l|l}
\toprule
 \textbf{Context Snippet}  & \multicolumn{1}{c|}{\textbf{Question}}& \multicolumn{1}{c}{\textbf{Gold Answer}}  \\ 
\midrule
\parbox{7cm}{\textcolor{red}{"Diamonds"} was certified sextuple platinum by the Recording Industry Association of America (RIAA). In Canada, \textcolor{red}{the song} debuted at number nine on the \textcolor{blue}{Canadian Hot 100} for the issue dated October 13, 2012 [...] \textcolor{red}{It} remained atop of \textcolor{blue}{it} for four consecutive weeks [...]}  & \parbox{5cm}{What is the full name of the chart of which Diamonds remained atop for four consecutive weeks?} & Canadian Hot 10\\
\midrule
\parbox{7cm}{The ever-winding path of \textcolor{blue}{John Frusciante}'s solo career is a confusing one to say the least [...] The album of the same name is \textcolor{blue}{Frusciante}'s first experimenting with the acid house genre. \textcolor{blue}{He} previously released an EP, Sect In Sgt under this alias in 2012.} & \parbox{5cm}{Who did release an EP called Sect In Sgt?} & John Frusciante \\
\bottomrule
\end{tabular}

\caption{Examples from our dataset. The context is cropped to only show the relevant parts.}
\label{table:chal_examples}
\end{table*}

\section{Creating an MRC Dataset that Better Reflects Coreference Reasoning}
\label{sect:challenge_set}
There is a growing trend in using adversarial models for data creation to make the dataset more challenging or discard examples that can be solved using surface cues \citep{Bartolo_2020,nie-etal-2020-adversarial,yang-etal-2018-hotpotqa,zellers-etal-2018-swag,yang2017mastering,dua-etal-2019-drop,chen2019codah,dasigi2019quoref}. 

Quoref is also created using an adversarial data collection method to discard examples that can be solved using simple lexical cues.
The assumption is that it is hard to avoid simple lexical cues by which the model can answer questions without coreference reasoning. Therefore, an adversarial model ($A$) is used to discard examples that contain such lexical cues.
While this adversarial filtering removes examples that are easy to solve by $A$, it does not ensure that the remaining examples do not contain shortcuts that are not explored by $A$.
First, the adversarial model in Quoref is trained on another dataset, i.e., SQuAD. Thus, the failure of $A$ on Quoref examples may be due to (1) Quoref having different lexical cues than those in SQuAD, or (2) domain shift.
Second, and more importantly, as argued by \citet{dunietz-etal-2020-test}, making the task challenging by focusing on examples that are more difficult for existing models is not a solution for more useful reading comprehension.\footnote{As put by them: ``the dominant MRC research paradigm is like trying to become a professional sprinter by glancing around the gym and adopting any exercises that look hard''.}
 
We instead propose a methodology for creating question-answer pairs as follows: 
\squishlist
	
	\item Annotators should create a question that connects the referring expression $m_1$ to its antecedent $m_2$ so that (1) $m_2$ is more informative than $m_1$,\footnote{Proper names are more informative than common nouns, and they are more informative than pronouns \citep{lee-etal-2013-deterministic}.} and (2) $m_1$ and $m_2$ reside in a different sentence.  %
    \item Candidate passages for creating QA pairs are selected according to their number of named entities and pronouns. The number of distinct named entities is an indicator of the number of entities in the text. Therefore, there would be more candidate entities for resolving referring expressions. The number of pronouns indicates that we have enough candidate $m_1$s that have more informative antecedents. 

\squishend

We provide this guideline to a student from the Computer Science department for generating new QA pairs from the existing passages in the Quoref development set.
We use Quoref passages to ensure that the source of performance differences on our dataset vs.\ Quoref is not due to domain differences.
This results in 200 new QA pairs. 
Table~\ref{table:chal_examples} presents examples from our dataset.

Table~\ref{table:quoref_challenge_bias_results} shows the results of the examined biases on our dataset.
By comparing Table~\ref{table:quoref_challenge_bias_results} and Table~\ref{table:bias_results}, we observe that the examined biases are less strong in our dataset, and their distribution is closer to those in CoNLL-2012. 
As we will see in Table~\ref{table:tase-results}, the performance of state-of-the-art models on Quoref drops more than 10 points, i.e., 13-18 points, on our challenge dataset.\footnote{We examine 50 randomly selected examples from our challenge set, and they were all answerable by a human.}

\begin{table}[!htb]
\footnotesize
\centering
\begin{tabular}{lc}
\toprule
 \bfseries Bias  & \bfseries Ours \\ 
\midrule
random named entity  & 3.03     \\
wh-word  & 13.64   \\ 
empty question  & 11.62   \\
semantic overlap & 24.50 \\
short-distance reasoning    & 35.35 \\
\bottomrule
\end{tabular}
\caption{Proportion of biased examples in our dataset.}
\label{table:quoref_challenge_bias_results}
\end{table} %
\section{Improving Coreference Reasoning}
\label{sect:method}
While we do not have access to many coreference annotations for the task of coreference-aware MRC, there are various datasets for the task of coreference resolution. 
Coreference resolution datasets contain the annotation of expressions that refer to the same entity.
In this paper, we hypothesize that we can directly use coreference resolution corpora to improve the coreference reasoning of MRC models.
We propose an effective approach to convert coreference annotations into QA pairs so that models learn to perform coreference resolution by answering those questions.
In our experiments, we use the CoNLL-2012 dataset \cite{pradhan-conll-2012} that is the largest annotated dataset with coreference information. %
\begin{table*}[!htb]
\footnotesize
\centering
 \resizebox{\textwidth}{!}{%
\begin{tabular}{l|c|l|l|c}
\toprule
 \textbf{Passage in CoNLL}  & \multicolumn{1}{c}{\textbf{Mention Cluster}}& \multicolumn{1}{|c}{\textbf{CoNLL$_{dec}$ Quesion}} & \multicolumn{1}{|c}{\textbf{CoNLL$_{bart}$ Question}} & \multicolumn{1}{|c}{\textbf{Gold Answer}}  \\ 
\midrule
\parbox{5cm}{\textcolor{blue}{My mother} was Thelma Wahl [...] \textcolor{blue}{She} was a very good mother. \textcolor{blue}{She} was at Huntingdon because \textcolor{blue}{she} needed care [...]}  & \textcolor{blue}{[My mother, She, She, she]} & \parbox{3cm}{She was at Huntingdon because $<$ref$>$ \textcolor{blue}{she} $<$/ref$>$ needed care.} &\parbox{3cm}{who was at huntingdon because she needed care?}  & My mother \\
\midrule
\parbox{5cm}{The angel also held \textcolor{blue}{a large chain} in his hand [...] The angel tied the dragon with \textcolor{blue}{the chain} for 1000 years.} & \textcolor{blue}{[a large chain, the chain]} & \parbox{3cm}{The angel tied the dragon with $<$ref$>$ \textcolor{blue}{the chain} $<$/ref$>$ for 1000 years.} & \parbox{3cm}{what did the angel tie the dragon with for 1000 years?} & a large chain \\
\bottomrule
\end{tabular}
 }
\caption{Coreference-to-QA conversion examples using CoNLL$_{dec}$ and CoNLL$_{bart}$ approaches.}
\label{tabel:qg_examples}
\end{table*}

\subsection{Coreference-to-QA Conversion} 
The existing approach to convert coreference annotations into (question, context, answer) tuples, which is used to improve coreference resolution performance \citep{wu-etal-2020-corefqa,aralikatte2020simple}, is to use the sentence of the anaphor as a declarative query, and its closest antecedent as the answer.
The format of these queries is not compatible with questions in MRC datasets, and therefore, the impact of this data on MRC models may be limited.
In this work, we instead generate questions from those declarative queries using an automatic question generation model. 
We use the BART model \cite{lewis-etal-2020-bart} that is one of the state-of-the-art text generation models.
Below we explain the details of each of these two approaches for creating QA data from CoNLL-2012.
Table~\ref{tabel:qg_examples} shows examples from both approaches. 
\paragraph{CoNLL$_{dec}$:} \citet{wu-etal-2020-corefqa} and \citet{aralikatte2020simple} choose a sentence that contains an anaphor as a declarative query, the closest non-pronominal antecedent of that anaphor as the answer, and the corresponding document of the expressions as the context.\footnote{We use the code provided by \citet{aralikatte2020simple}.} 
	We remove the tuples in which the anaphor and its antecedent are identical. The reason is that (1) Quoref already contains many examples in which the coreference relation is between two mentions with the same string, and (2) even after removing such examples, CoNLL$_{dec}$ contains around four times more QA pairs than the Quoref training data.
\paragraph{CoNLL$_{bart}$:} we use a fine-tuned BART model \cite{lewis-etal-2020-bart} released by \citet{durmus-etal-2020-feqa} for question generation and apply it on the declarative queries in CoNLL$_{dec}$. The BART model specifies potential answers by masking noun phrases or named entities in the query and then generates questions for each masked text span. We only keep questions whose answer, i.e., the masked expression, is a coreferring expression and replace that answer with its closest non-pronominal antecedent. We only keep questions in which the masked expression and its antecedent are not identical.  
Such QA pairs enforce the model to resolve the coreference relation between the two coreferring expressions to answer generated questions.

\subsection{Experimental Setups}
We use two recent models from the Quoref leaderboard: RoBERTa \citep{liu2019roberta} and TASE \cite{segal2020simple}, from which TASE has the state-of-the-art results.
We use RoBERTa-large from HuggingFace \citep{wolf2020huggingfaces}.
TASE casts MRC as a sequence tagging problem to handle questions with multi-span answers. 
It assigns a tag to every token of the context indicating whether the token is a part of the answer. 
We use the TASE$_{IO}$+SSE setup that is a combination of their multi-span architecture and single-span extraction with IO tagging.%
We use the same configuration and hyper-parameters for TASE$_{IO}$+SSE as described in \citet{segal2020simple}.
We train all models for two epochs in all experiments.\footnote{The only difference of TASE in our experiments and the reported results in \citet{segal2020simple} is the number of training epochs. For a fair comparison, we train all models for the same number of iterations.}
We use the F1 score that calculates the number of shared words between predictions and gold answers for evaluation.
\paragraph{Training Strategies.}
To include the additional training data that we create from CoNLL-2012 using coreference-to-CoNLL conversion methods, we use two different strategies:
\squishlist
\item \emph{Joint}: we concatenate the training examples from Quoref and CoNLL-to-QA converted datasets. Therefore, the model is jointly trained on the examples from both datasets. 
\item \emph{Transfer}: Since the CoNLL-to-QA data is automatically created and is noisy, we also examine a sequential fine-tuning setting in which we first train the model on the CoNLL-to-QA converted data, and then fine-tune it on Quoref.
\squishend

\subsection{Data}
\label{sec:data}

\begin{table*}[!htb]
\footnotesize
\centering
\begin{tabular}{llccccc}
\toprule
 \textbf{Model} & \textbf{Training setup}  & \multicolumn{1}{c}{\textbf{Quoref$_{dev}$}} & \multicolumn{1}{c}{\textbf{Quoref$_{test}$}} & \multicolumn{1}{c}{\textbf{Ours}} & \multicolumn{1}{c}{\textbf{Contrast set}} & \multicolumn{1}{c}{\textbf{MultiRC}} \\ %
\midrule
\multirow{6}{*}{TASE}
& Baseline & 84.05 & 84.71	& 66.48	&73.44	&51.83 \\ %
& CoNLL$_{bart}$ & \textcolor{gray}{34.95} & \textcolor{gray}{35.76}	& \textcolor{gray}{39.55}	& \textcolor{gray}{26.24}	& \textcolor{gray}{26.51}	\\ %
& Joint-CoNLL$_{dec}$    & 84.36 & 85.14	& \textcolor{gray}{65.92}	& 74.88	& \textcolor{gray}{44.71}  \\ %
& Transfer-CoNLL$_{dec}$  & 85.00 & 85.88	& \bfseries 73.07	& 75.69	& \textcolor{gray}{50.18} 	\\ %
& Joint-CoNLL$_{bart}$    & 84.30 & 85.93 &	69.37 & 74.00	& \textcolor{gray}{48.26} \\ %
& Transfer-CoNLL$_{bart}$ & \bfseries 85.13 &  85.98	& 73.01	&  77.40 &  51.96  \\ %
& Transfer-SQuAD & 84.70 & \bfseries 87.02 & 67.99 & \bfseries 78.28 & \bfseries 53.51  \\
\midrule
\multirow{6}{*}{RoBERTa}
&  Baseline &   {79.64}	&  79.69  & 64.35	& \bfseries 69.95	&  37.12	\\ %
& CoNLL$_{bart}$ & \textcolor{gray}{28.82} & \textcolor{gray}{29.10} & \textcolor{gray}{29.00}	& \textcolor{gray}{17.36}	& \textcolor{gray}{14.81}	\\ %
& Joint-CoNLL$_{dec}$ & \textcolor{gray}{75.15} & \textcolor{gray}{74.83} & \textcolor{gray}{56.94} & \textcolor{gray}{57.78}	& \textcolor{gray}{29.97} \\ %
& Transfer-CoNLL$_{dec}$  & \textcolor{gray}{74.10} & \textcolor{gray}{73.65} & \textcolor{gray}{60.09}	& \textcolor{gray}{58.95}	& \textcolor{gray}{30.93} 	\\ %
& Joint-CoNLL$_{bart}$ & \textcolor{gray}{78.70} & \textcolor{gray}{79.59} & \bfseries 67.07 	& \textcolor{gray}{66.78}	& \textcolor{gray}{35.43}  \\ %
& Transfer-CoNLL$_{bart}$ & \textcolor{gray}{78.22} & \textcolor{gray}{78.33} & {66.62}	& \textcolor{gray}{66.58}	& \textcolor{gray}{36.84}  \\ %
& Transfer-SQuAD & \bfseries 80.18 & \bfseries 79.82 & 64.88 & \textcolor{gray}{69.46} & \bfseries 38.26 \\
\bottomrule
\end{tabular}
\caption{Impact of incorporating coreference data in MRC using CoNLL$_{dec}$ and CoNLL$_{bart}$ conversion methods on RoBERTa-large and TASE models.
The \emph{Baseline} and \emph{CoNLL$_{bart}$} rows show the results when models are trained on the Quoref training data and the CoNLL$_{bart}$ data, respectively. 
 \emph{Joint} refers to the setting in which the model is jointly trained on Quoref and the converted CoNLL data. \emph{Transfer} refers to the setting in which the model is first trained on the converted CoNLL data and fine-tuned on Quoref.  \emph{Transfer-SQuAD} shows the impact of training the model on additional QA data from a similar domain. 
 Results are reported based on F1 scores. The highest F1 scores for each model are boldfaced and scores lower than the \emph{Baseline} are marked in gray. }
\label{table:tase-results}
\end{table*}

We evaluate all the models on four different QA datasets. 
\begin{itemize}
	\item \emph{Quoref}: the official development and test sets of Quoref, i.e., Quoref$_{dev}$ and Quoref$_{test}$, respectively. 
	\item \emph{Our challenge set}: our new evaluation set described in Section~\ref{sect:challenge_set}. 
		\item \emph{Contrast set}: the evaluation set by \citet{gardner2020evaluating} that is created based on the official Quoref test set. For creating this evaluation set, the authors manually performed small but meaningful perturbations to the test examples in a way that it changes the gold label. This dataset is constructed to evaluate whether models’ decision boundaries align to true decision boundaries when they are measured around the same point. %

	\item \emph{MultiRC}: Multi-Sentence Reading Comprehension set \cite{MultiRC2018} is created in a way that answering questions requires a more complex understanding from multiple sentences. 
	Therefore, coreference reasoning can be one of the sources for improving the performance on this dataset. 
	Note that \emph{MultiRC} is from a different domain than the rest of evaluation sets.\footnote{To use the MultiRC development set, which is in a multi-choice answer selection format, we convert it to a reading comprehension format by removing QA pairs whose answers cannot be extracted from the context.}

\end{itemize}

The \emph{Contrast set} and \emph{MultiRC} datasets are not designed to explicitly evaluate coreference reasoning. However, we include them among our evaluation sets to have a broader view about the impact of using our coreference data in QA.

\begin{table}[!htb]
\footnotesize
\centering
\begin{tabular}{lr|lr}
\toprule
 \textbf{Training}  & \multicolumn{1}{c|}{\textbf{examples}} & \textbf{Test}  & \multicolumn{1}{c}{\textbf{examples}}  \\ 
\midrule
Quoref train  & 19399 &  Quoref dev & 2418 \\
CoNLL$_{dec}$ & 89403  & Ours & 200 \\
CoNLL$_{bart}$ & 18906 & Contrast set & 700  \\
SQuAD & 86588 & MultiRC & 389  \\

\bottomrule
\end{tabular}
\caption{Number of examples in each dataset.}
\label{table:data}
\end{table}

Table~\ref{table:data} reports the statistics of these QA datasets.
In addition, it reports the number of examples in CoNLL$_{dec}$ and CoNLL$_{bart}$ datasets that we create by converting the CoNLL-2012 training data into QA examples.
Since the question generation model cannot generate
a standard question for every declarative sentence, CoNLL$_{bart}$ contains a smaller number of examples. 
We also include the statistics of SQuAD in Table \ref{table:data} as we use it for investigating whether the resulting performance changes are due to using more training data or using coreference-aware additional data.

The language of all the datasets is English.

\subsection{Results}
Table~\ref{table:tase-results} presents the results of evaluating the impact of using coreference annotations to improve coreference reasoning in MRC. 
We report the results for both of the examined state-of-the-art models, i.e., TASE and RoBERTa-large, using both training settings: (1) training the model jointly on Quoref and CoNLL-to-QA converted 
data (Joint), and (2) pre-training the model on CoNLL-to-QA data
first and fine-tuning it on Quoref (Transfer).
\emph{Baseline} represents the results of the examined models that are only trained on Quoref.
\emph{CoNLL$_{bart}$} represents the results of the models that are only trained on the CoNLL$_{bart}$ data.
\emph{Transfer-SQuAD} reports the results of the sequential training when the model is first trained on the SQuAD training dataset \citep{rajpurkar-etal-2016-squad} and is then fine-tuned on Quoref.

Based on the results of Table~\ref{table:tase-results}, we make the following observations.

First, 
the most successful setting for improving coreference reasoning, i.e., improving the performance on our challenge evaluation set, is Transfer-CoNLL$_{bart}$. 
Pre-training the TASE model on \emph{CoNLL$_{bart}$} improves its performance on all of the examined evaluation sets.
However, it only improves the performance of RoBERTa on our challenge set. 

\begin{table*}[!htb]
\footnotesize
\centering
\begin{tabular}{lcccc|cccc}
\toprule
\bfseries Model &\multicolumn{2}{c}{\textbf{Semantic overlap}} 
&\multicolumn{2}{c|}{\textbf{$\neg$Semantic overlap}} &\multicolumn{2}{c}{\textbf{Short reasoning}} &\multicolumn{2}{c}{\textbf{$\neg$Short reasoning}} \\
\midrule
& dev & Ours & dev & Ours & dev & Ours & dev & Ours \\
\midrule
TASE Baseline                & 81.69  & 77.2         & 84.86  & 62.96        & 94.84  & 89.04        & 72.95  & 54.15 \\ \hline
Joint-CoNLL$_{dec}$     &  +2.07  & \textcolor{gray}{-5.80}         & \textcolor{gray}{-0.30}   & +1.19        & +0.94  & \textcolor{gray}{-1.65}        & \textcolor{gray}{-0.34}  & +0.03 \\
Joint-CoNLL$_{bart}$    & +0.86  & \textcolor{gray}{-3.00}           & +0.03  & +4.82        & +0.64  & +1.20         & \textcolor{gray}{-0.16}  & +3.80  \\
Transfer-CoNLL$_{dec}$  & +1.29  &  +8.56        & +0.83  & +5.94        & +1.07  &  +8.82        &  +0.83  & +5.37 \\
Transfer-CoNLL$_{bart}$ & +1.74  & +1.23        &  +0.85  &  +8.26        &  +1.54  & +4.70         & +0.60   &  +7.51 \\
Transfer-SQuAD & +0.84	& +0.58	&  +1.19	& +0.10 & \textcolor{gray}{-0.91}	& +2.3 & +0.33	& +2.15 \\
\midrule 
RoBERTa baseline & 78.09	& 67.39	& 80.19	& 63.36	& 90.04	& 84.23	& 68.97	& 53.48  \\ \hline
Joint-CoNLL$_{dec}$ & \textcolor{gray}{-5.55}	& \textcolor{gray}{-10.04}	& \textcolor{gray}{-4.15}	& \textcolor{gray}{-6.55}	& \textcolor{gray}{-2.53}	& \textcolor{gray}{-7.32}	& \textcolor{gray}{-6.54}	& \textcolor{gray}{-7.46} \\
Joint-CoNLL$_{bart}$ & 0.00  &  +1.94	& \textcolor{gray}{-1.28}	& +2.97	& \textcolor{gray}{-0.48}	&  \textcolor{gray}{-1.36}	& \textcolor{gray}{-1.43}	& +4.95 \\
Transfer-CoNLL$_{dec}$ & \textcolor{gray}{-4.20}	& +1.79	& \textcolor{gray}{-6.02}	& \textcolor{gray}{-6.27}	& \textcolor{gray}{-3.52}	& \textcolor{gray}{-7.00}	& \textcolor{gray}{-7.65}	& \textcolor{gray}{-2.77} \\
Transfer-CoNLL$_{bart}$ & \textcolor{gray}{-1.02}	& \textcolor{gray}{-0.55}	& \textcolor{gray}{-1.58}	&  +3.18	& \textcolor{gray}{-0.95}	& \textcolor{gray}{-5.06}	& \textcolor{gray}{-1.94}	&  +6.27 \\
Transfer-SQuAD &  +1.32	& \textcolor{gray}{-1.08}	&  +0.25	& +1.05	&  +0.45	& \textcolor{gray}{-9.46}	&  +0.6	& +5.99 \\
\bottomrule
\end{tabular}
\caption{F1 score differences of various TASE and RoBERTa models on the Quoref$_{dev}$ and our dataset splits that are created based on the semantic overlap and short distance reasoning biases. For instance, \emph{Ours} in the \emph{$\neg$Semantic overlap} column shows the performance differences of the examined models on the split of our dataset in which examples do not contain the semantic overlap bias. Negative differences are marked in gray.}
\label{table:lex_results}
\end{table*} 

Second, %
SQuAD contains well-formed QA pairs while CoNLL$_{bart}$ and CoNLL$_{dec}$ contain noisy QA. Also, SQuAD and Quoref are both created based on Wikipedia articles, and therefore, have similar domains. However, the genres of the documents in CoNLL-2012 include newswire, broadcast news, broadcast conversations, telephone conversations, weblogs, magazines, and Bible, which are very different from those in Quoref.
As a result, pretraining on SQuAD has a positive impact on the majority of datasets. However, this impact is less pronounced on our challenge dataset, as it requires coreference reasoning while this skill is not present in SQuAD examples.

Finally, while using the sentence of coreferring mentions as a declarative query (CONLL$_{dec}$) is the common method for converting coreference resolution datasets into QA format in previous studies, our results show using CoNLL$_{bart}$ has a more positive impact compared to using CoNLL$_{dec}$. 

\subsection{Analysis}
To analyze what kind of examples benefit more from incorporating the coreference data, we split Quoref$_{dev}$ and our dataset into different subsets based on the \emph{semantic overlap} and \emph{short distance reasoning} biases, which are the most common types of biases in both datasets.

The \emph{semantic overlap} column in Table~\ref{table:lex_results} represents the results on the subset of the data in which answers reside in the most similar sentence of the context, and the \emph{$\neg$semantic overlap} column contains the rest of the examples in each of the examined datasets.
The \emph{short reasoning} column presents the results on the subset of the data containing examples that can be solved by the short distance reasoning bias model, and \emph{$\neg$ short reasoning} presents the results on the rest of the examples.

Table~\ref{table:lex_results} shows the performance differences of the TASE and RoBERTa models on these four subsets for each of the two datasets.

Surprisingly, the performance of the baseline models is lower on the \emph{semantic overlap} subset compared to \emph{$\neg$semantic overlap} on Quoref$_{dev}$. This can indicate that examples in the \emph{$\neg$semantic overlap} subset of Quoref$_{dev}$ contain other types of biases that make QA less challenging on this subset.

The addition of the coreference resolution annotations in all four training settings reduces the performance gap of the TASE model on the \emph{semantic overlap} and \emph{$\neg$semantic overlap} subsets for both datasets. 
Incorporating coreference data for RoBERTa, on the other hand, has a positive impact using the CoNLL$_{bart}$ data and on the harder subsets of our challenge evaluation set, i.e., \emph{$\neg$semantic overlap} and \emph{$\neg$short reasoning}.

Finally, 
there is still a large performance gap between \emph{short reasoning} and \emph{$\neg$ short reasoning} subsets. In our coreference-to-QA conversion methods, we consider the closest antecedent of each anaphor as the answer. A promising direction for future work is to also create QA pairs based on longer distance coreferring expressions, e.g., to create two QA pairs based on each anaphor, one in which the answer is the closest antecedent, and the other with the first mention of the entity in the text as the answer.

\section{Conclusions}
We show that the high performance of recent models on the Quoref dataset
does not necessarily indicate that they are adept at performing coreference reasoning,
and that QA based on coreference reasoning is a greater challenge than current scores suggest.
We then propose a methodology for creating a dataset that better presents the coreference reasoning challenge for MRC.
We provide our methodology to an annotator and create a sample dataset.
Our analysis shows that our dataset contains fewer biases compared to Quoref, and the performance of state-of-the-art Quoref models drops considerably on this evaluation set.

To improve the coreference reasoning of QA models, we propose to use coreference resolution datasets to train MRC models.
We propose a method to convert coreference annotations into an MRC format.
We examine the impact of incorporating this coreference data on improving the coreference reasoning of QA models using two top-performing QA systems from the Quoref leaderboard.
We show that using coreference datasets improves the performance of both examined models on our evaluation set, indicating their improved coreference reasoning.
The results on our evaluation set suggest that there is still room for improvement, and reading comprehension with coreference understanding remains a challenge for existing QA models, especially if the coreference relation is between two distant expressions.

\section*{Acknowledgments}
This work has been supported by the German Research Foundation (DFG) as part of the QASciInf project (grant GU 798/18-3), and the German Federal Ministry of Education and Research and the Hessian Ministry of Higher Education, Research, Science and the Arts within their joint support of the National Research Center for Applied Cybersecurity ATHENE.
Dan Roth's work is partly supported by contract FA8750-19-2-1004 with the US Defense Advanced Research Projects Agency (DARPA). 
The authors would like to thank Michael Bugert, Max Glockner, Yevgeniy Puzikov, Nils Reimers, Andreas R{\"u}ckl{\'e}, and the anonymous reviewers for their valuable feedback. 
\bibliographystyle{acl_natbib}
\interlinepenalty=10000
\bibliography{acl2020,anthology}

\appendix
\section{Additional Statistics about Biased Examples}
Table ~\ref{table:conll_dec_bias_results} shows the proportion of biased examples in the CoNLL$_{dec}$ set. We can see that the results are similar to that of the CoNLL$_{bart}$ set. 

\begin{figure*}[!htb]
\centering
\includegraphics[width=0.75\paperwidth]{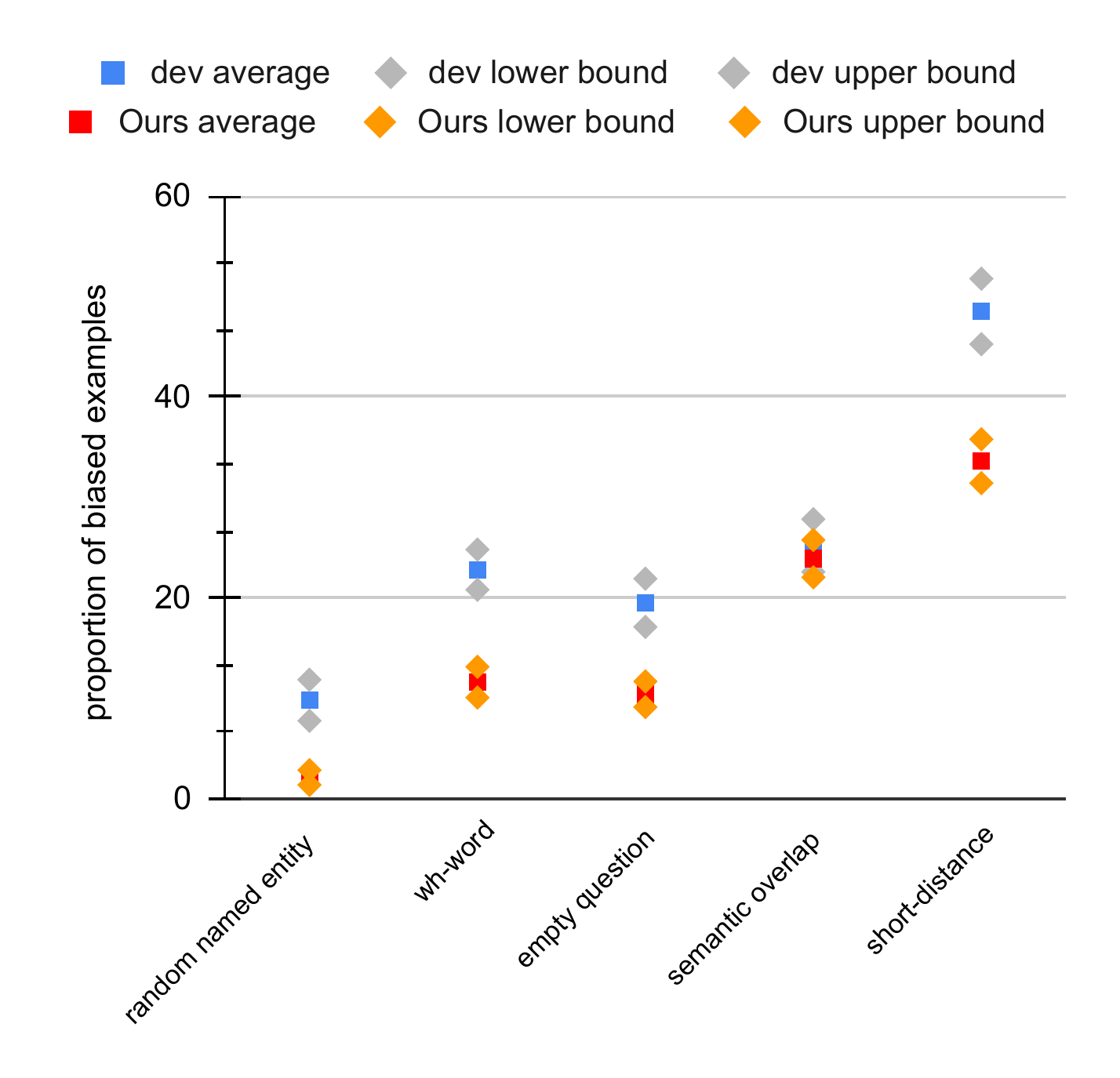}
\caption{The average, upper and lower bounds of the ratio of biased examples in Quoref$_{dev}$ and our challenge set for the randomly sampled 10 subsets.}
\label{bias_statistic}
\end{figure*}

To compare the ratio of biased examples between Quoref$_{dev}$ and our challenge set when considering the same number of examples in both datasets, we randomly sample 10 different subsets from Quoref$_{dev}$ and our challenge set with 100 samples in each subset and compute the rations in each subset. %
Figure~\ref{bias_statistic} shows the results. As we see, in this setting the ratio of all bias types in our evaluation set is still lower than those in Quoref$_{dev}$. 

\section{Additional Experiments}
Table~\ref{table:additional_results} shows additional experiments for pre-training the examined models on coreference data.
We examine an additional setting for pre-training on both CoNLL$_{dec}$ and CoNLL$_{bart}$ by first training the models on CoNLL$_{dec}$, then on CoNLL$_{bart}$, and finally on Quoref (\emph{Transfer-CoNLL$_{bart+dec}$}).
By comparing the results of \emph{Transfer-CoNLL$_{bart+dec}$} with \emph{Transfer-CoNLL$_{bart}$} from Table~\ref{table:tase-results}, we observe that pre-training the models on both CoNLL$_{dec}$ and CoNLL$_{bart}$ does not result in any additional advantage compared to only using CoNLL$_{bart}$. 

\section{Additional Examples}
Table~\ref{tabel:qg_examples_more} presents more examples from CoNLL$_{dec}$ and CoNLL$_{bart}$.

\begin{table}[!htb]
\footnotesize
\centering
\begin{tabular}{lc}
\toprule
 \bfseries Bias  & \bfseries CoNLL$_{dec}$ \\ 
\midrule
random named entity  & 2.11    \\
wh-word  & 12.97   \\ 
empty question  & 11.24   \\
semantic overlap & 35.32 \\
short-distance reasoning    & 21.05 \\

\bottomrule
\end{tabular}
\caption{Proportion of biased examples in CoNLL$_{dec}$ dataset.}
\label{table:conll_dec_bias_results}
\end{table}

\begin{table*}[!htb]
\footnotesize
\centering
\begin{tabular}{llccccc}
\toprule
 \textbf{Model} & \textbf{Training setup}  & \multicolumn{1}{c}{\textbf{Quoref$_{dev}$}} & \multicolumn{1}{c}{\textbf{Quoref$_{test}$}} & \multicolumn{1}{c}{\textbf{Ours}} & \multicolumn{1}{c}{\textbf{Contrast set}} & \multicolumn{1}{c}{\textbf{MultiRC}} \\
\midrule
\multirow{2}{*}{TASE}
& Baseline & 84.05 & 84.71	& 66.48	&73.44	&51.83 \\ 
& Transfer-CoNLL$_{bart+dec}$ & 85.01 & 85.73 & 68.06 & 76.54 & 49.61 \\
\midrule
\multirow{2}{*}{RoBERTa}
&  Baseline &   {79.64}	&  79.69  & 64.35	& 69.95	&  37.12	\\
& Transfer-CoNLL$_{bart+dec}$ & 73.29  & 73.73  & 58.19	& 57.18	&  31.50 \\ 
\bottomrule
\end{tabular}
\caption{Additional experiments on using the CoNLL$_{dec}$ and CoNLL$_{bart}$ data for pre-training RoBERTa-large and TASE models. \emph{Transfer-CoNLL$_{bart+dec}$} refers to the setting in which the model is first trained on CoNLL$_{dec}$, then on CoNLL$_{bart}$, and finally on Quoref. 
}
\label{table:additional_results}
\end{table*}

\begin{table*}[!htb]
\footnotesize
\centering
 \resizebox{\textwidth}{!}{%
\begin{tabular}{l|c|l|l|c}
\toprule
 \textbf{Passage in CoNLL}  & \multicolumn{1}{c}{\textbf{Mention Cluster}}& \multicolumn{1}{|c}{\textbf{CoNLL$_{dec}$ Quesion}} & \multicolumn{1}{|c}{\textbf{CoNLL$_{bart}$ Question}} & \multicolumn{1}{|c}{\textbf{Gold Answer}}  \\ 
\midrule
\parbox{5cm}{After George W. Bush is sworn in, \textcolor{blue}{Bill Clinton} will head to New York. \textcolor{blue}{Mr. Clinton} will also spend time at \textcolor{blue}{his} presidential library in Arkansas. \textcolor{blue}{He} says \textcolor{blue}{he} will come to Washington, 'every now and then'.} & \textcolor{blue}{[Bill Clinton, Mr. Clinton, his, He, he]} & \parbox{3cm}{He says $<$ref$>$ \textcolor{blue}{he} $<$/ref$>$ will come to Washington, `every now and then.'} & \parbox{3cm}{who says he will come to washington, 'every now and then'?} & Bill Clinton \\
\midrule
\parbox{5cm}{\textcolor{blue}{Paul} had already decided not to stop at Ephesus. \textcolor{blue}{He} did not want to stay too long in Asia. \textcolor{blue}{He} was hurrying because \textcolor{blue}{he} wanted to be in Jerusalem on the day of Pentecost if possible.} & \textcolor{blue}{[Paul, He, He, he]}  & \parbox{3cm}{He was hurrying because $<$ref$>$ \textcolor{blue}{he} $<$/ref$>$  wanted to be in Jerusalem on the day of Pentecost if possible.} &\parbox{3cm}{who was hurrying because they wanted to be in jerusalem on the day of pentecost if possible?}  & Paul  \\
\midrule 
\parbox{5cm}{\textcolor{blue}{The KMT vice chairman} arrived at party headquarters to meet with KMT Chairman Lien Chan on the afternoon of pw...\textcolor{blue}{He} said that \textcolor{blue}{he} will follow Lien Chan as a lifelong volunteer.} & \textcolor{blue}{[The KMT vice chairman, He, he]}  & \parbox{3cm}{He said that $<$ref$>$ \textcolor{blue}{he} $<$/ref$>$ will follow Lien Chan as a lifelong volunteer.} &\parbox{3cm}{who said that he will follow lien chan as a lifelong volunteer?}  & The KMT vice chairman  \\
\midrule 
\parbox{5cm}{...It also includes a lot of sheep, good clean - living, healthy sheep, and \textcolor{blue}{an Italian entrepreneur} has an idea about how to make a little money of them...So \textcolor{blue}{this guy} came up with the idea of having people adopting sheep by an internet.} & \textcolor{blue}{[an Italian entrepreneur, this guy]}  & \parbox{3cm}{So $<$ref$>$ \textcolor{blue}{this guy} $<$/ref$>$ came up with the idea of having people adopting sheep by an internet.} & \parbox{3cm}{who came up with the idea of having people adopting sheep by an internet?} & an Italian entrepreneur \\
\midrule 
\parbox{5cm}{George W. Bush has met with \textcolor{blue}{Al Gore} in Washington. The two men met for just 15 minutes at the Vice President's official residence...Bush went into the talks with \textcolor{blue}{his defeated rival} after meeting with President Clinton earlier today.} & \textcolor{blue}{[Al Gore, his defeated rival]} & \parbox{3cm}{Bush went into the talks with $<$ref$>$ \textcolor{blue}{his defeated rival} $<$/ref$>$ after meeting with President Clinton earlier today.} &\parbox{3cm}{who did bush go into the talks with after meeting with president clinton earlier today?} & Al Gore  \\
\midrule 
\parbox{5cm}{Meanwhile \textcolor{blue}{Prime Minister Ehud Barak} told Israeli television he doubts a peace deal can be reached before Israel's February 6th election. \textcolor{blue}{He} said \textcolor{blue}{he} will now focus on suppressing Palestinian violence.} & \textcolor{blue}{[Prime Minister Ehud Barak, He, he]} &\parbox{3cm}{He said $<$ref$>$ \textcolor{blue}{he} $<$/ref$>$ will now focus on suppressing Palestinian violence.} &\parbox{3cm}{who said he will now focus on suppressing palestinian violence?} & Prime Minister Ehud Barak \\
\bottomrule
\end{tabular}
 }
\caption{More examples from coreference-to-QA conversion using CoNLL$_{dec}$ and CoNLL$_{bart}$ approaches.}
\label{tabel:qg_examples_more}
\end{table*}

\end{document}